\documentclass[a4paper]{article}

\usepackage{INTERSPEECH2016}

\usepackage{graphicx}
\usepackage{amssymb,bm}
\usepackage{textcomp}
\usepackage[tbtags]{amsmath}
\usepackage[utf8]{inputenc}
\usepackage{cite}

\def\vec#1{\ensuremath{\bm{{#1}}}}
\def\mat#1{\vec{#1}}

\DeclareMathOperator*{\argmin}{argmin}
\newcommand {\X} {\mat{X}}
\newcommand {\V} {\mat{V}}
\newcommand {\Y} {\mat{Y}}
\newcommand {\Xhat} {\mat{\hat{X}}}
\newcommand {\Vhat} {\mat{\hat{V}}}
\newcommand {\Yhat} {\mat{\hat{Y}}}
\newcommand {\Ux} {\mat{U}_X}
\newcommand {\Uy} {\mat{U}_Y}
\newcommand {\x} {\vec{x}}
\newcommand {\xhat} {\vec{\hat{x}}}
\newcommand {\fTheta} {\hat{f}_{\Theta}}
\newcommand {\summation} {\sum\limits}

\ninept

\title{Dictionary Update for NMF-based Voice Conversion Using an Encoder-Decoder Network}

\makeatletter
\def\name#1{\gdef\@name{#1\\}}
\makeatother
\name{{\em Chin-Cheng Hsu$^1$, Hsin-Te Hwang$^1$, Yi-Chiao Wu$^1$, Yu Tsao$^2$, and Hsin-Min Wang$^1$}}

\address{
  $^1$Institute of Information Science, Academia Sinica, Taipei, Taiwan \\
  $^2$Research Center for Information Technology Innovation, Academia Sinica, Taipei, Taiwan \\
  {\small \tt \{jeremycchsu, hwanght, tedwu, whm\}@iis.sinica.edu.tw, yu.tsao@citi.sinica.edu.tw}
}

\begin{document}

\maketitle

\begin{abstract}
 In this paper, we propose a dictionary update method for Non-negative Matrix Factorization (NMF) with high dimensional data in a spectral conversion (SC) task. 
 Voice conversion has been widely studied due to its potential applications such as personalized speech synthesis and speech enhancement. 
 Exemplar-based NMF (ENMF) emerges as an effective and probably the simplest choice among all techniques for SC, as long as a source-target parallel speech corpus is given. 
 ENMF-based SC systems usually need a large amount of bases (exemplars) to ensure the quality of the converted speech. However, a small and effective dictionary is desirable but hard to obtain via dictionary update, in particular when high-dimensional features such as STRAIGHT spectra are used. 
 Therefore, we propose a dictionary update framework for NMF by means of an encoder-decoder reformulation. 
 Regarding NMF as an encoder-decoder network makes it possible to exploit the whole parallel corpus more effectively and efficiently when applied to SC. 
 Our experiments demonstrate significant gains of the proposed system with small dictionaries over conventional ENMF-based systems with dictionaries of same or much larger size.
\end{abstract}

\section{Introduction}
  The purpose of voice conversion is to transform spectral and prosodic characteristics of an utterance from a source speaker so that the perceived speaker identity matches a target speaker, with other information, such as the linguistic contents, unaltered. In this study, we focus on spectral conversion (SC), whereas inspection on prosody conversion is beyond the scope of this paper.

\begin{figure}[t]
    \includegraphics[width=0.45\textwidth]{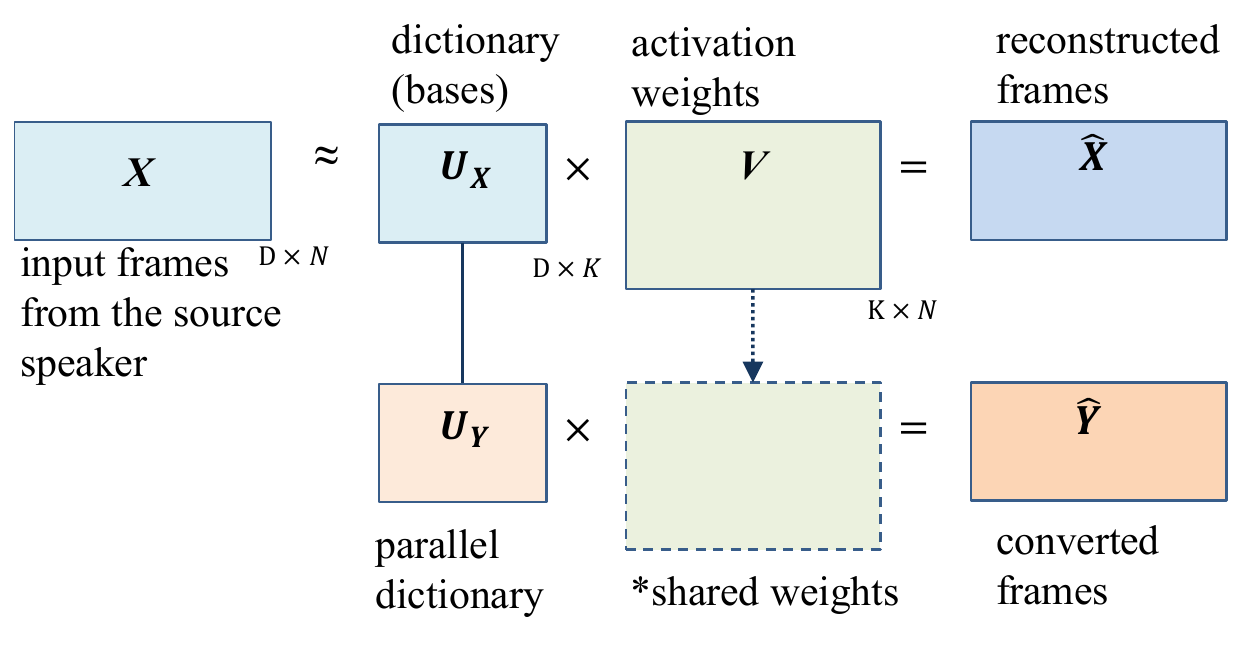}
    \centering
    \caption{\it{Basic idea of ENMF-based spectral conversion.}}
    \label{fig:NMF-VC}
  \end{figure}

  A wide variety of techniques have been applied to SC, including Gaussian mixture models (GMMs)
  \cite{Toda07, GMMTakamichi, Stylianou98},
  frequency warping
  \cite{Erro10-FreqWarp, Godoy12-FreqWarp},
  deep neural networks (DNNs)
  \cite{Desai10-NN-VC, Nakashika13-NN-VC, Hwang15-NN-VC, Chen14-NN-VC},
  and exemplar-based approaches
  \cite{Takashima12-Exemplar, Wu13-ENMF-VC, Wu14-NMF-VC, Wu16-LLE-VC}.
 Among them, exemplar-based non-negative matrix factorization (ENMF), a confluence of exemplar-based approaches and NMFs \cite{Lee01-NMF}, is considered exceptionally suitable for the task of SC \cite{Wu14-NMF-VC}. An ENMF-based SC system has a meaningful set of basis frames (a.k.a. dictionary) that reconstructs a given input. Conversion is achieved simply via applying the activation weights of the source dictionary to the target dictionary. The paired source and target dictionaries are constructed beforehand from a parallel corpus. Systems of this kind can be applied on the fly (without training procedures). Figure \ref{fig:NMF-VC} depicts how ENMF-based SC is conducted.

  Aside from the advantages, ENMF-based SC approaches have several practical shortcomings. 
  First, it is unclear how dictionary update could be applied to improve the pair of source and target dictionaries. 
  Second, multiplicative update becomes costly when the dimensionality and number of samples become huge. 
  Third, one still needs a mechanism for dictionary selection in order to reduce the conversion time, although the robustness of ENMF-based SC increases with the dictionary size. 

Attempts have been made to tackle the above problems. For example, the authors of \cite{Aihara16-SemiNMF} relaxed the non-negativity constraint for the dictionaries and adopted features with lower dimension (mel-cepstra coefficients) to make computation manageable. In addition, they dropped the assumption of weight sharing during dictionary update but instead posed a similarity constraint on the individual source and target activation weights so that the paired dictionaries could be learned separately. 

In this paper, we propose a reformulation of NMF as an encoder-decoder network (EDN) and conduct dictionary learning using mini-batch Stochastic Gradient Descent (SGD). With this formulation, we are able to obtain compact dictionaries that can well reconstruct the source and target frames, respectively. Our experiments demonstrate that, even with much smaller dictionaries, the proposed EDN-based SC system outperforms the conventional ENMF-based SC system. 
  
  The rest of this paper is organized as follows. We first briefly review ENMF-based spectral conversion in Section \ref{ENMF-SC}. Then, we describe our proposed method in Section \ref{OurMethod}, and present our experimental results and discussions in Section \ref{Experiments}. Finally, Section \ref{Conclusion} concludes this paper.
  

\section{ENMF-based spectral conversion}
\label{ENMF-SC}
  NMF factorizes a sequence of spectral frames $\X=[\x_1, \x_2, ..., \x_N]$ 
  into a dictionary matrix $\Ux=[\vec{u}_1, \vec{u}_2, ..., \vec{u}_K]$ 
  and an activation weight matrix $\V=[\vec{v}_1, \vec{v}_2, ..., \vec{v}_N]$:  
  \begin{equation}
  \label{eq:ENMF-X}
    \X \approx \Ux \V.
  \end{equation}
Typically, $\Ux$ and $\V$ can be learned alternately by keeping the other matrix fixed. 

For applying NMF to SC, there are one requirement and one assumption. 
First, a set of parallel frames (source-target frame pairs) whose acoustic contents are aligned is required. 
The paired source and target dictionaries, $\Ux$ and $\Uy$, can therefore be directly constructed from the exemplars. 
Second, the intrinsic topologies of the source and target speech have to share some similarity. 
Satisfying these two conditions reduces SC to a self-reconstruction problem, letting ENMF come into play. 
Similarity of intrinsic topologies implies similarity of the activation weights for reconstruction. 
In this manner, conversion is realized by finding the shared activation weights $\V$ from the source speech $\X$ and then applying them directly to the target dictionary $\Uy$ to generate the converted speech $\Y$:
  \begin{equation}
  \label{eq:ENMF-V}
    \V^*=\argmin_{\V} D(\Ux \V, \X) + C(\V),
  \end{equation}
  \begin{equation}
  \label{eq:conversion}
    \Y \approx \Uy \V,
  \end{equation}  
  where $D(\cdot, \cdot)$ is a cost function in terms of, for example, mean square error, and $C(\cdot)$ is a constraint such as L1 norm (sparsity). It should be noted that there are no training phases in ENMF-based SC. Conversion, i.e., solving $\V$ based on (\ref{eq:ENMF-V}) and then applying the resulting $\V^*$ to (\ref{eq:conversion}), is conducted online.

  \section{The proposed encoder-decoder-based dictionary update framework}
  \label{OurMethod}
  \subsection{NMF as an encoder-decoder network}
    To better understand the proposed encoder-decoder framework for SC, we begin with reformulating ENMF as an encoder-decoder pair. 
    We will then generalize it to accommodate NMF in SC.
    Here, ENMF is distinguished from NMF as follows. 
    If the dictionary is directly derived from exemplars without training, it is called ENMF; otherwise, it is NMF.
    In ENMF, the goal is to find the best activation $\V$ given the input $\X$ and dictionary $\Ux$. We can reformulate ENMF in (\ref{eq:ENMF-V}) and (\ref{eq:ENMF-X}) as
    \begin{equation}
      \V=f(\X, \Ux),
    \end{equation}
    \begin{equation}
    \label{X-approx-UV}
      \X \approx g_X(\V, \Ux )=\Ux \V.
    \end{equation}
    Although $f(\cdot)$ has no analytic form due to the non-negativity constraint, we can still approximate it using a feed-forward neural network $\fTheta(\cdot)$ parameterized by $\Theta$ as
    \begin{equation}
      \V \approx \Vhat=\fTheta (\X).
    \end{equation}
    We refer to $\fTheta(\cdot)$ as the encoding function (a.k.a. encoder). Function $g_X(\cdot)$ in (\ref{X-approx-UV}) naturally emerges as a decoding function (a.k.a. decoder) because it reconstructs the input using the code $\V$. The encoder-decoder pair ($\fTheta$, $g_X$) thus constitutes an auto-encoder. With  $\fTheta(\cdot)$ designed to yield non-negative outputs (e.g. having rectified outputs), we can approximate an ENMF using an auto-encoder \cite{Smaragdis15-NMF-eq-NN}.

      The decoder $g_X (\cdot)$ is parameterized by $\Ux$, which can be trained, and training $\Ux$ corresponds to dictionary update in NMF terminology. Dictionary update brings performance gain at the cost of requiring a training process, which is absent in ENMF. 
      As both the activation and the dictionary are learnable, the auto-encoder now approximates an NMF. Taking NMF as an auto-encoder gives us a clearer insight on how to update the parallel dictionaries for SC, which will be described in \ref{sec:dict-update}.

    \subsection{Encoder-decoder network for spectral conversion}
\begin{figure}[t]
        \includegraphics[width=0.35\textwidth]{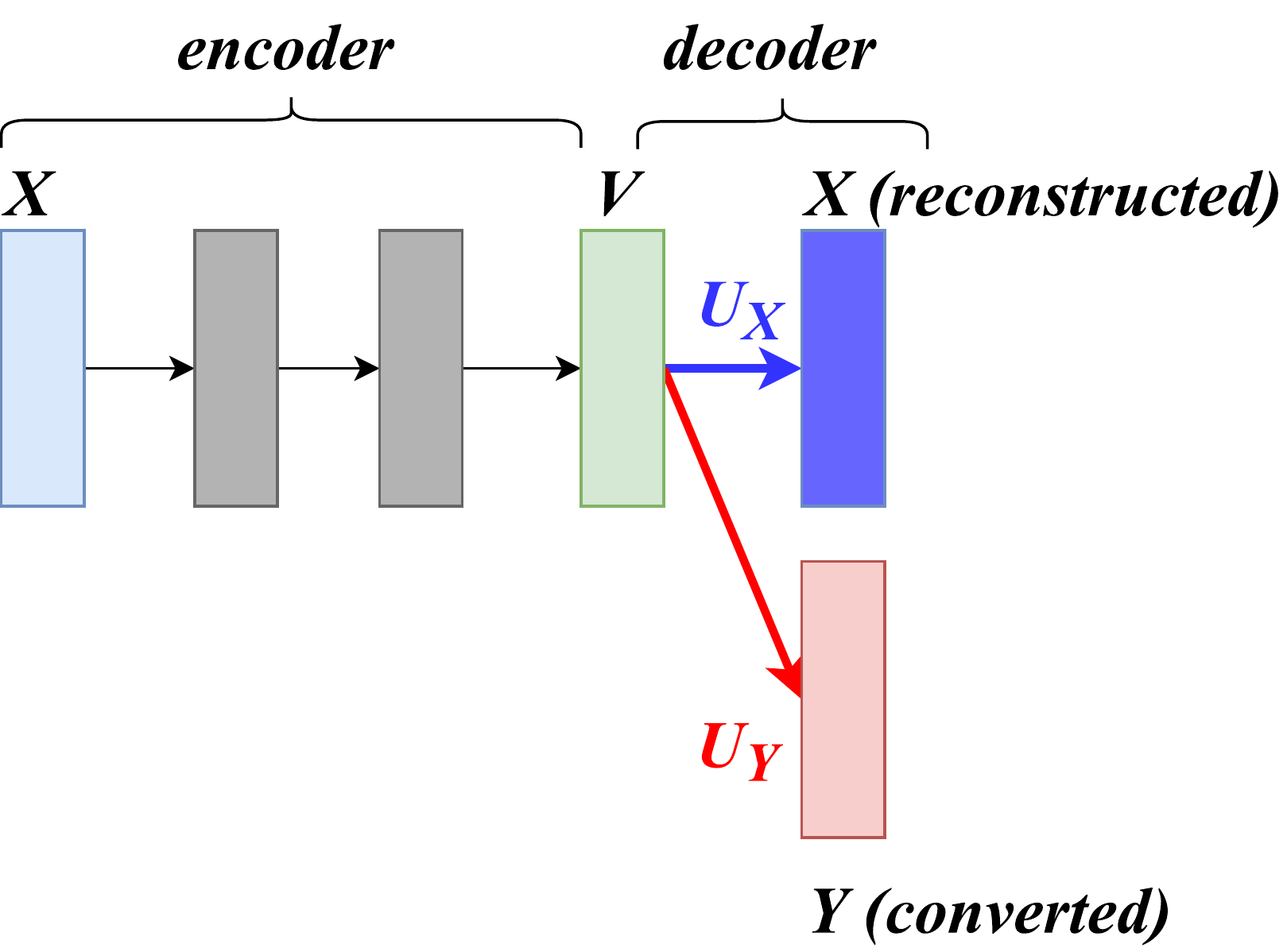}
        \centering
        \caption{\it{Illustration of EDN-based spectral conversion. There is only one encoder because only source speech is available during the conversion phase.}}
        \label{fig:Architecture}
      \end{figure}

      Now we extend the encoder-decoder framework to tackle the problem of SC.
      Conversion requires an additional decoder that transforms the code $\V$ into a target spectrum without changing linguistic contents. This is exactly what the target dictionary $\Uy$ is for. We can again regard it as a decoder, denoted by $g_Y(\cdot)$, and concatenate it to the encoder $\fTheta(\cdot)$. There are one encoder and two juxtaposed decoders now. For brevity, we refer to the proposed method as an Encoder-Decoder Network (EDN). Figure \ref{fig:Architecture} depicts its architecture. We summarize it using the following equations:
      \begin{equation}\label{eq:encode-to-v}
        \text{Encoding}: \fTheta(\X)=\Vhat,
      \end{equation}
      \begin{equation}\label{eq:decode-to-x}
        \text{Decoding to the source}: g_X(\Vhat)=\Ux \Vhat=\Xhat,
      \end{equation}
      \begin{equation}\label{eq:decode-to-y}
        \text{Decoding to the target}: g_Y(\Vhat)=\Uy \Vhat=\Yhat.
      \end{equation}
      Note that the decoders merely perform a linear transformation with either dictionary $\Ux$ or $\Uy$.
      The conversion phase is similar to that of the ENMF-base SC. Spectral frames $\X$ from a source speaker are fed into the encoder network to get the code $\Vhat$, which is then used by the target decoder to obtain the target frames $\Yhat$ (cf. (\ref{eq:encode-to-v}) and (\ref{eq:decode-to-y})).
      
    \subsection{Training the EDN}
    \subsubsection{Training the encoder}
      The training process of EDN is divided into two stages. The first stage involves only the auto-encoder, which approximates an ENMF. The encoder parameters $\Theta$ are updated to minimize the approximation divergence between input $\X$ and the auto-encoder output $\Xhat$, which is defined as,
      \begin{equation}
      \label{eq:encoder}
         J_{encoder}
        = D(\X, \Xhat) 
        = \summation_{n=1}^{N} D_{KL}(\x_n||\xhat_n),
      \end{equation}
      where $\x_n$ is a reference frame (ground truth), 
      $\xhat_n$ is its reconstructed version, 
      $n$ is the frame index, 
      and
      $N$ is the number of frames.
      Kullback-Leibler divergence (KLD), which is defined as
      \begin{equation}
      \label{KLD}
      D_{KL}(\x_n || \xhat_n) 
      := \summation_{m=1}^M {x}_{m, n} \frac{\log{{x}_{m, n}}}{\log{\hat{x}}_{m, n}},
      \end{equation}
      where $m$ is the dimension index, 
      and $M$ is the dimensionality, 
      is chosen as the cost function $D(\cdot, \cdot)$ for the following reasons. 
      When dealing with power or magnitude spectra, the order of magnitude varies across dimensions drastically. 
      Logarithms in KLD help alleviate the issue. 
      In our experiments, both input and output spectra are normalized to unit-sum so that KLD can be applied readily.

      Substituting $\Xhat$ in (\ref{eq:encoder}) with (\ref{eq:decode-to-x}) and then $\Vhat$ with (\ref{eq:encode-to-v}), the cost function becomes
      \begin{equation}
      \label{eq:encoder-ext}
      \begin{split}
        J_{encoder}
        = & D(\X, \Ux \Vhat) \\
        = & D(\X, \Ux \fTheta(\X)) \\
        = & \summation_{n=1}^N D_{KL}(\x_n || \Ux \fTheta(\x_n)).
      \end{split}
      \end{equation}
      Keeping $\Ux$ fixed, we update the encoder parameters by
      \begin{equation}
        \Theta^*=\argmin_{\Theta} J_{encoder}.
      \end{equation}

      Encoder training is very similar to the root finding procedures of ENMF-based SC (cf. (\ref{eq:ENMF-V})), except that we do not use the multiplicative update rules. Mini-batches of source frames are fed into the encoder, gradients are computed, and $\Theta$ is updated using the SGD rules. 

    \subsubsection{Training the decoders}
    \label{sec:dict-update}
      The second stage of the training process of EDN is for dictionary update, or equivalently, decoder training.
      The decoder parameters ($\Ux, \Uy$) are initially the parallel dictionaries, which consist of randomly selected exemplars, used in ENMF-based SC. Their ability to reconstruct or convert speech is limited, and that is why we need to update them. Similar to (\ref{eq:encoder}) and (\ref{eq:encoder-ext}), we first define a joint cost function for the two decoders:
      \begin{equation}
      \begin{split}
      \label{eq:alpha}
         J_{decoder} 
        = & \summation_{n=1}^N 
          \alpha D(\X, \Xhat) + (1 - \alpha) D(\Y, \Yhat) \\
        = & \summation_{n=1}^N 
          \alpha D_{KL}(\x_n || \Ux \fTheta(\x_n)) \\
        + & \summation_{n=1}^N (1 - \alpha) D_{KL}(\vec{y}_n || \Uy \fTheta(\x_n)),
      \end{split}
      \end{equation}
      where $\alpha$ is the importance weight of self-reconstruction. We also choose KLD in (\ref{KLD}) as our cost functions. Dictionaries are updated by
      \begin{equation}
      \{\Ux^*, \Uy^*, \Theta^*\} = \argmin_{\Ux, \Uy, \Theta} {J_{decoder}}.
      \end{equation}

      Special care should be taken of the dictionaries. 
      We actually apply a rectifier and a unit-sum normalizer to them so that the resulting dictionaries conform to our specified forms. 
      Note that we also update the encoder in the second stage. 

      We conduct training as follows. The EDN takes as input a frame {$\x_n$} from the source speaker, and tries to minimize the divergence between the decoder outputs ($\xhat_n, \vec{\hat{y}}_n$) and the regression targets ($\x_n$, $\vec{y}_n$), which are the input itself and its corresponding frame from the target. 
      Similar procedures are applied to update the encoder parameters $\Theta$.
      Note that, to avoid notation cluttering, we describe the procedures using \emph{frame} pairs as opposed to \emph{mini-batch} pairs that are used in practice.
            
      This architecture allows us to update dictionaries $\Ux$ and $\Uy$ separately in form, jointly in reality since each paired source and target frames share the same activation weights. This multi-task design avoids updating a joint parallel dictionary with twice the original dimension.

  \section{Experiments}
  \label{Experiments}
    \subsection{Experimental settings}
      \subsubsection{The VCC2016 speech corpus}
        The proposed SC system was evaluated on a parallel English corpus from the Voice Conversion Challenge 2016 \cite{VCC2016}. 
        There are 5 male and 5 female speakers in this corpus. 
        Each speaker has 150 utterances as the training set and 12 utterances as the evaluation set. 
        Five out of the ten speakers are designated to be the conversion targets (2 female and 3 male speakers) and the other five sources (3 female and 2 male speakers). 
        The testing set comprises 54 utterances per target speaker.

       We conducted experiments on a subset of the speakers. 
       Two speakers were chosen as sources (SF1 and SM1) and another two as targets (TF2 and TM3). 
       We reported two types of spectral conversion: intra-gender and cross-gender.

      \subsubsection{Feature sets}
        We used the STRAIGHT toolkit \cite{Kawahara99-STRAIGHT} to parametrize speech into the smoothed STRAIGHT spectra (SP for short), aperiodicity (AP), and pitch contours (F0). The FFT length was set to 1024, so the resulting AP and SP were both 513-dimensional. The frame shift was 5 milliseconds (ms) and the frame length was 25 ms. Neither contextual nor dynamic features were utilized in any forms. The SP was converted using our proposed method or the baseline systems. All systems converted F0 using the same linear mean-variance transformation. Energy and AP were kept unmodified. In all the systems, every input frame of SP was normalized to unit-sum. 
        After spectral conversion, energy was compensated back to SP, and STRAIGHT took in all the parameters to synthesize utterances.

        Each parallel training set was aligned using dynamic time warping (DTW) with 24-ordered Mel-cepstral coefficients (MCC) extracted from SP. Energy-based voice activity detection (VAD) was used to exclude the silence segments. After alignment, the length of a source utterance remained the same while some frames from the target were duplicated or decimated.

    \subsection{The baseline systems}
    The first baseline SC system was a conventional ENMF-based one (as described in \cite{Wu14-NMF-VC}) with dictionaries of 512 bases. For each source-target pair, 512 randomly selected frames from the whole source training samples were specified as the source dictionary $\Ux$. Their corresponding frames from the target speaker were specified to be the target dictionary $\Uy$. The system is denoted as ENMF-512 in the following experiments. An extended baseline (ENMF-3000) was built on larger dictionaries of 3000 bases (the previously selected 512 bases plus additional 2488 bases). The unit-sum constraint naturally poses a sparse constraint on the baseline systems.

      \subsection{The encoder-decoder network-based system}
      \subsubsection{Configurations and hyper-parameters}
        The encoder was a feed-forward neural network with 2 hidden layers, each with 1024 nodes.
        Rectifier linear unit (ReLU) \cite{Nair10-ReLU} was applied to each layer to provide non-linearity and to ensure the non-negativity constraint of the activation.
        The batch size was 512.
        Both the activation and the dictionaries were re-normalized to unit-sum.
        Note that the dictionaries are outputs from ReLUs followed by a unit-sum normalization, so the non-negativity can be guaranteed.

        The maximum number of epochs for encoder training was set to 100, and the learning rate was 0.001. As for dictionary update, the learning rate was set to 0.01 and decreased by a factor of 0.1, for three times per 50 epochs. The optimizers were Adam \cite{Kingma15-Adam}. Usually it only took a few dozens of epochs to reach a reasonable solution. The importance coefficient $\alpha$ in (\ref{eq:alpha}) was empirically set to 0.15, striking a balance between the reconstruction and the conversion divergence. It might be tempting to set $\alpha$ to 0 so that EDN could focus on conversion only. However, our preliminary results revealed no obvious advantages for such setting, presumably because of imperfect frame alignments.


      \subsubsection{Training procedures}
        For each source-target pair, the 512-base source and target dictionaries used in the baseline ENMF-512 system served as initial $\Ux$ and $\Uy$, respectively.
        As mentioned in Sec. \ref{OurMethod}, the encoder parameters $\Theta$ were trained in the first stage of EDN training, and then the dictionaries $\Ux$ and $\Uy$ and the encoder parameters $\Theta$ were updated in the second stage of EDN training.
        We denote our proposed EDN-based SC system with 512 bases as EDN-512 in the following experiments.

      \begin{figure}[t]
        \includegraphics[width=0.45\textwidth]{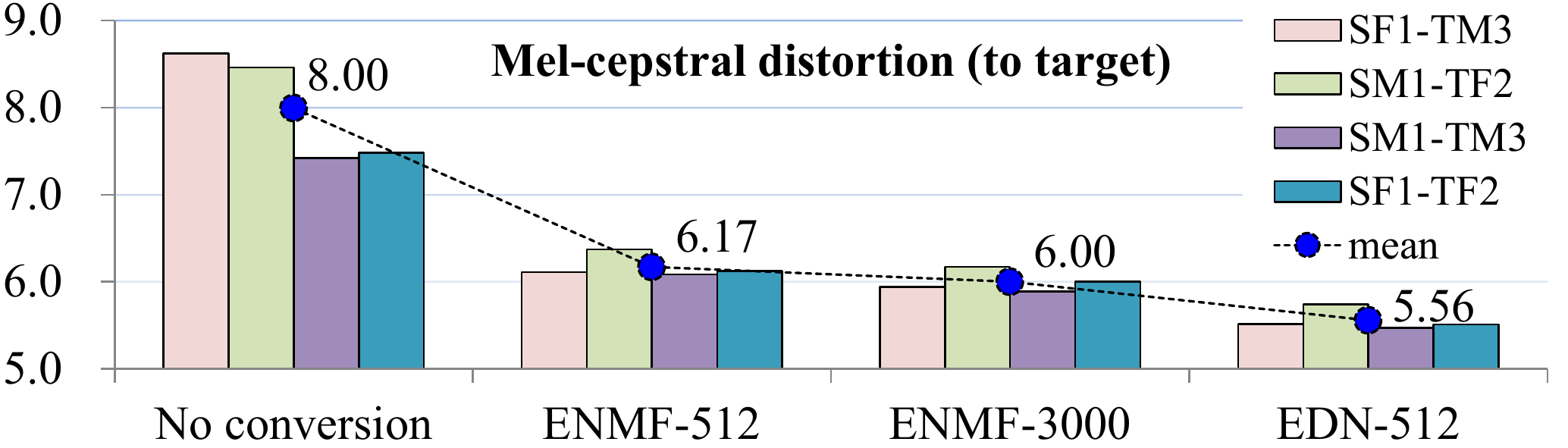}
        \centering
        \caption{\it{Graphical comparison of MCDs.}}
        \label{fig:MCD}
      \end{figure}

    \subsection{Objective evaluation}
    \subsubsection{Mel-cepstral distortion}
      We visualize mean mel-cepstral distortion (MCD) values on the evaluation set in Figure \ref{fig:MCD}. Our proposed method (EDN-512) achieved the lowest distortion under all test conditions.

      Performance gain of EDN-512 over ENMF-512 is attributed to training, which grants EDN access to the whole training set.
      Compared to ENMF-3000, EDN-512 is superior in that it achieves better quality with fewer bases. 
      This fact indicates the effectiveness of the training process, which brings EDN a stronger power of representation so that the code can be decoded into the target speech more precisely.

    \subsubsection{Code sparsity}

\begin{figure}[t]
        \includegraphics[width=0.4\textwidth]{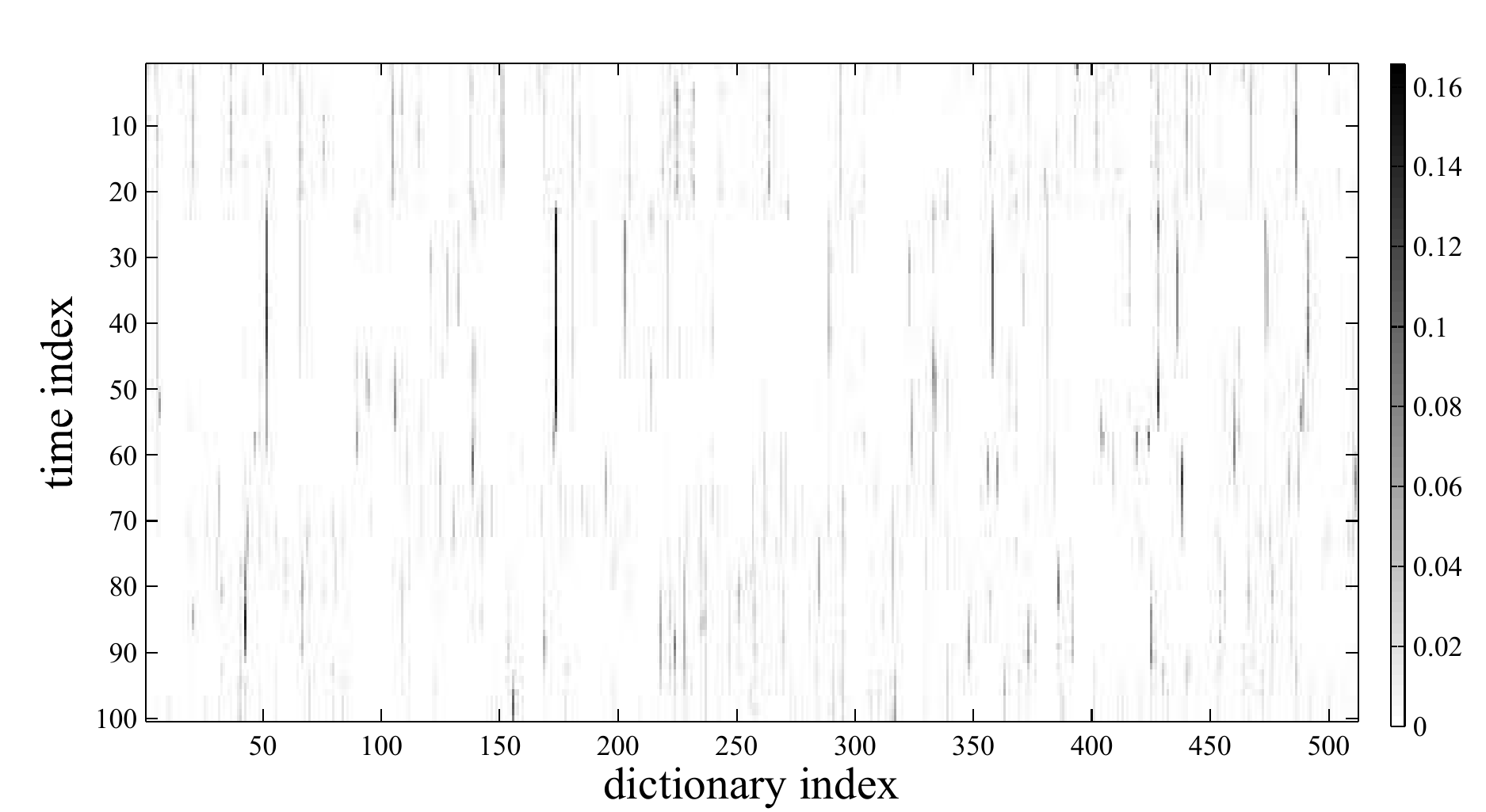}
        \centering
        \caption{\it{EDN activation (code) matrix of a typical voiced speech segment. Most entries are exactly zero, meaning that the code is highly sparse.}}
        \label{fig:Code}
      \end{figure}

	We can observe three things from the resulting activation (code) $\V$ shown in Figure \ref{fig:Code}.
	First, the activation is still sparse after dictionary update, implicitely proving that the EDN simulates an NMF.
	Most of the activation weights are exactly zero thanks to ReLU non-linearity. For typical voiced frames (e.g. from the 20th to the 100th) , usually less than 100 bases are activated, and only a few of them dominate.
	Second, consecutive frames have similar activations, meaning that temporal smoothness is guaranteed.
	Third, the fact that the codes are shared indicates that they carry certain speaker-independent information (assumably, acoustics) because they can be decoded into voices of different speakers.

    \subsection{Subjective evaluation}
\begin{figure}[t]
        \includegraphics[width=0.375\textwidth]{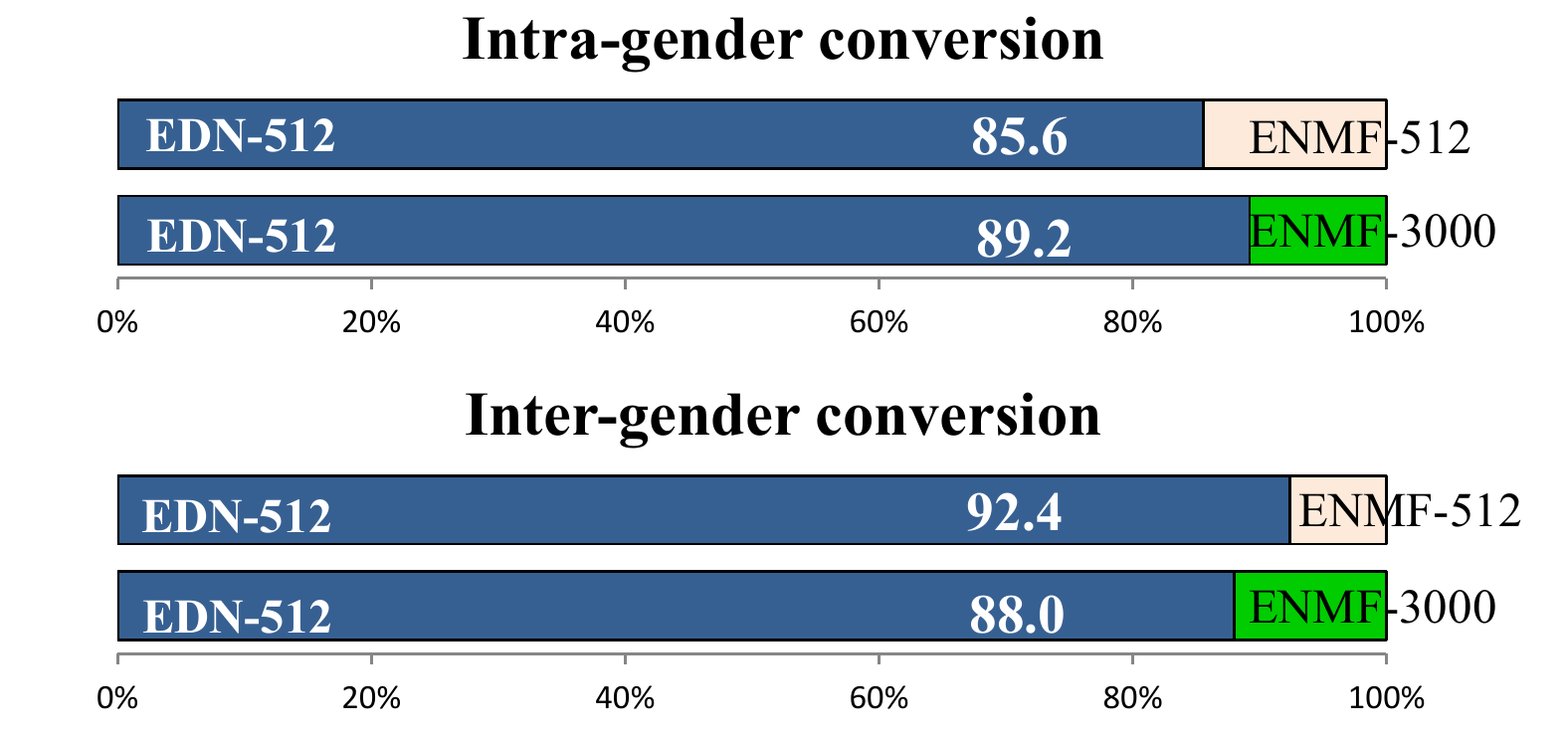}
        \centering
        \caption{\it{Preference score from the ABX test for conversion quality. The source is a female speaker (SF1), the intra-gender target is TF2, and the inter-gender target is TM3.}}
        \label{fig:ABX}
      \end{figure}

      We also evaluated voice quality using the ABX test. Ten listeners were invited to evaluate all the testing sets, each with 25 sentences. The results are shown in Figure \ref{fig:ABX}. Similarity was not reported because all the three methods achieved a nearly identical level of similarity. 

      Obviously, subjective evaluation demonstrated significantly higher voice quality of the proposed EDN-based system over that of the two ENMF-based baseline systems. The results are consistent with the objective evaluation.
      
  \section{Conclusions}
  \label{Conclusion}
    This paper has presented a dictionary update framework for NMF-based spectral conversion by reformulating NMF as an encoder-decoder network. The merits are two-fold. First, the proposed method avoids explicit joint dictionary update that doubles the dimensionality. Second, the learned dictionary is much more compact and has a higher representational power, resulting in better voice quality in the converted speech. The encoder-decoder network formulation can be easily generalized to application cases without the non-negativity constraint. We will consider these cases in the future.

  \newpage
  \eightpt
  \bibliographystyle{IEEEtran}
  \bibliography{ednbib}

\end{document}